\documentclass{article}
\usepackage{ijcai19}

\usepackage{times}
\usepackage{soul}
\usepackage{url}
\usepackage[hidelinks]{hyperref}
\usepackage[utf8]{inputenc}
\usepackage[small]{caption}
\usepackage{graphicx}
\usepackage{amsmath}

\usepackage{amssymb}
\usepackage{booktabs}
\usepackage{algorithm}
\usepackage{multirow}
\usepackage{paralist}
\usepackage{algorithmic}
\usepackage{xcolor}
\usepackage{bm}
\newcommand{\blue}[1]{{#1}}

\title{Adversarial Imitation Learning from Incomplete Demonstrations}

\author{
Mingfei Sun\footnote{Contact author}
\And
Xiaojuan Ma
\affiliations
Department of Computer Science and Engineering, Hong Kong University of Science and Technology
\emails
mingfei.sun@ust.hk,
mxj@cse.ust.hk
}

\begin{document}

\maketitle

\begin{abstract}
Imitation learning targets deriving a mapping from states to actions, a.k.a. \textit{policy}, from expert demonstrations. Existing methods for imitation learning typically require any actions in the demonstrations to be fully available, which is hard to ensure in real applications. 
% algorithms, e.g., Behavior Cloning and Inverse Reinforcement Learning, requires the actions 
% in demonstrations to be always available and noisy-free. 
Though algorithms for learning with unobservable actions have been proposed, they focus solely on state information and overlook the fact that the action sequence could still be partially available and provide useful information for policy deriving. In this paper, we propose a novel algorithm called Action-Guided Adversarial Imitation Learning (AGAIL) that learns a policy from demonstrations with incomplete action sequences, i.e., incomplete demonstrations. The core idea of AGAIL is to separate demonstrations into state and action trajectories, and train a policy with state trajectories while using actions as auxiliary information to guide the training whenever applicable. Built upon the \textit{Generative Adversarial Imitation Learning}, AGAIL has three components: a generator, a discriminator, and a guide. The generator learns a policy with rewards provided by the discriminator, which tries to distinguish state distributions between demonstrations and samples generated by the policy. The guide provides additional rewards to the generator when demonstrated actions for specific states are available. We compare AGAIL to other methods on benchmark tasks and show that AGAIL consistently delivers comparable performance to the state-of-the-art methods even when the action sequence in demonstrations is only partially available. 
% AGAIL also consistently outperforms all other algorithms when the actions are presented in different level of being noisy

% Imitation learning targets at finding a mapping from states to actions from demonstrations. Most existing algorithms require the demonstrations to be complete (with both states and actions available) and noise-free regardless of the fact the actions may not be available or observable and the demonstrated state-action pairs could be noisy. To overcome this problem, we propose Action Constrained Adversarial Imitation Learning (ACAIL), which aims to learn an expert policy under different action conditions, i.e., no action, partial actions, noisy actions, and noisy partial actions. \notice{how algorithm works}. We present theoretical analysis to prove the correctness of ACAIL. Experiments show that ACAIL outperforms most of the imitation learning algorithms, including the state-of-the-art algorithm (GAIL) when the actions are partial and noisy. \notice{detailed results}
\end{abstract}

\section{Introduction}
Imitation learning is a framework for learning a behavior policy from demonstrations. Usually, demonstrations are presented in the form of state-action trajectories, with each pair indicating the action to take at the state being visited. 
% The task is to learn a generalized action-taking strategy for all possible states that may be encountered in one task. 
In order to learn the behavior policy, the demonstrated actions are usually utilized in two ways. The first, known as Behavior Cloning (BC)~\cite{bain1999framework}, treats the action as the target label for each state, and then learns a generalized mapping from states to actions in a supervised manner~\cite{pomerleau1991efficient}. 
% In most conventional approaches to imitation learning, the demonstrations are given in the form of state-action trajectories, and, accordingly, the action in demonstrations often play a vital role There are two ways of using As an important component of demonstrations, the action has two important characteristics that significantly influences the performance of both the BC and IRL algorithms. 
% Imitation learning is a method to learn an optimal mapping between states and actions from expert demonstrations, which are given in the form of state-action trajectories, i.e., $\tau = [(s_1, a_1), (s_2, a_2),..., (s_t, a_t)]$. 
% Most imitation learning algorithms assume that the demonstrations  , with both states and actions being complete and noise-free~\cite{gao2018reinforcement}. Under this assumption, demonstrations are then utilized in two ways.
% Under this learning scenario, demonstrations are presented for the agent to imitate, and thus play an important role as they contain the optimal mapping from states to actions, a.k.a. \textit{policy}, how to reproduce the demonstrated behavior. 
% state-action pair is treated separately
% the whole sequence of state-action pairs are modeled
% Though working well for several practical applications~\cite{bojarski2016end,daftry2016learning}, BC usually suffers from \textit{cascading errors}~\cite{ross2010efficient}, a severe problem caused by minor inaccuracies compounding over time which lead to states that are dramatically different from what encountered in training.
% inverse reinforcement learning
Another way, known as Inverse Reinforcement Learning (IRL)~\cite{ng2000algorithms}, views the demonstrated actions as a sequence of decisions, and aims at finding a reward/cost function under which the demonstrated decisions are optimal. Once the reward/cost function is found, the policy could then be obtained through a standard Reinforcement Learning algorithm. 

\begin{figure}
    \centering
    \includegraphics[width=1.0\linewidth]{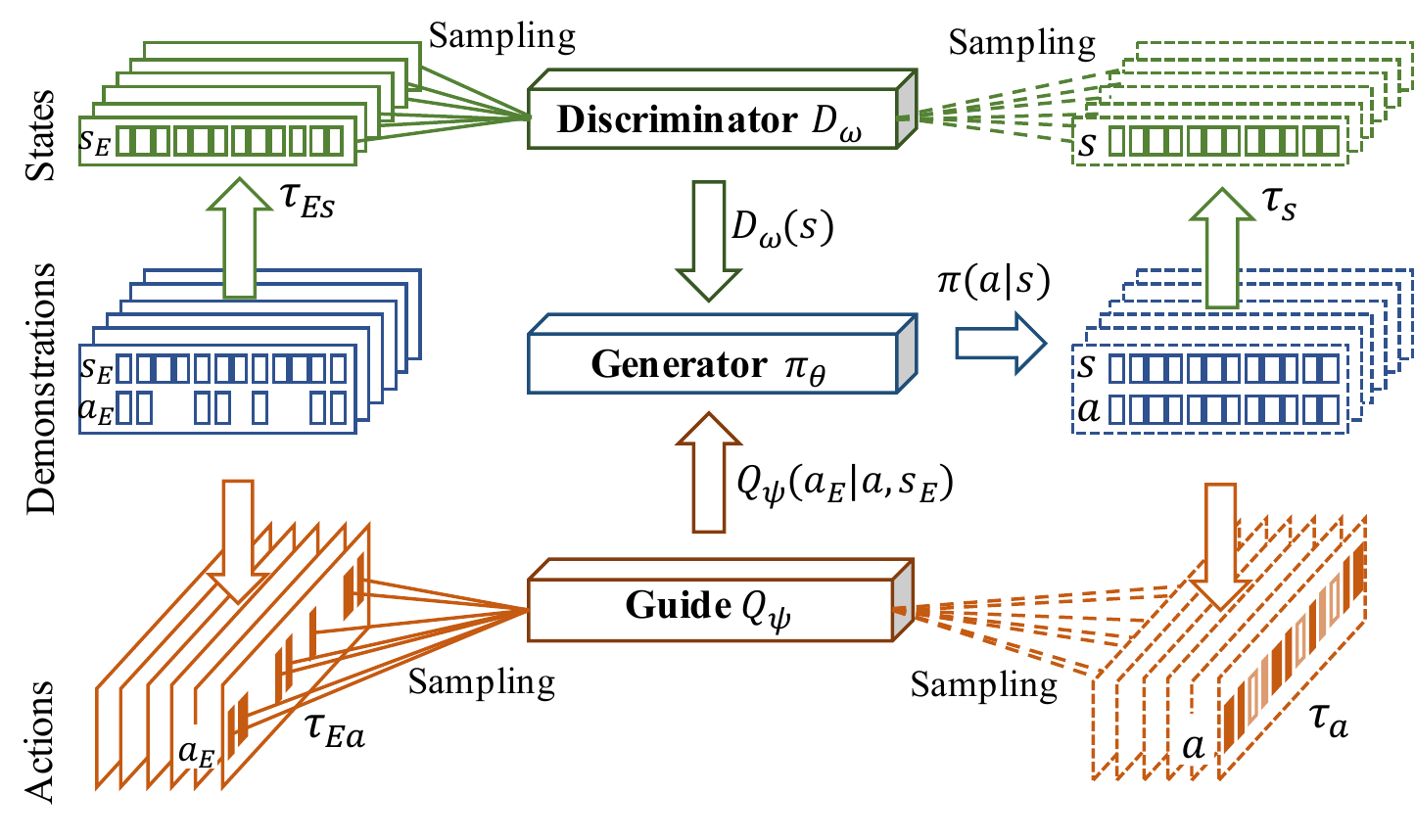}
    \caption{Action-Guided Adversarial Imitation Learning has three components: a generator, a discriminator, and a guide. The discriminator distinguishes state distributions between demonstrations and samples generated by the generator, i.e., policy. The guide provides auxiliary rewards to the generator whenever actions are available.}
    \label{fig:framework}
\end{figure}

Nevertheless, both BC and IRL algorithms implicitly assume that the demonstrations are complete, meaning that the action for each demonstrated state is fully observable and available ~\cite{gao2018reinforcement}. This assumption hardly holds for a real imitation learning task. First, the actions (not the states) in demonstrations may be partially observable or even unobservable~\cite{torabi2018behavioral}. For example, when showing a robot how to correctly lift up a cup, the demonstrator's states -- body movements -- can be visually captured but the human actions -- the force and torque applied to the body joint -- are unavailable to the robot~\cite{eysenbach2018diversity}.
Furthermore, even if the actions are obtainable, some of them may be invalid and need to be eliminated from learning due to the demonstrator's individual factors~\cite{argall2009survey}, e.g., the expertise level or strategy preferences~\cite{li2017infogail}
% while some other actions could still provide a certain level of guidance for learning.
% Even the same individual might make different actions while encountering the same situations, potentially resulting in demonstrations generated from multiple near-optimal but distinct policies. 
Without complete action information in demonstrations, the conventional BC and IRL algorithms are unable to produce the desired policy. 
% Furthermore, the policy trained with the noisy actions by BC algorithms could cause compound errors~\cite{ross2011reduction}, and, even if trained with IRL algorithms, the action noises could cause the learned policy to deviate from expert behavior~\cite{brys2015reinforcement}. 

% Without explicit action information in demonstrations, the conventional BC and IRL algorithms are unable to produce the desired policy.
% Even with a force sensor on the cup, the actions are only partially sensed since the force and torque of other body joint are unsensible. Generally, actions that do not affect the environment are not visible to an outside observer. 
% Furthermore, demonstrations without action information are quite common. For example, there is a great number of tutorial videos on YouTube that only provide the observer knowledge of the demonstrator's state trajectory. 
% \red{Instead of assuming that the demonstrations are complete and fully observable, we argue that the demonstrations are not needed all the time. For example, in autonomous driving, the expert could only make demonstrations when he/she thinks it is necessary.}

Though some recent studies have proposed to use state trajectories~\cite{merel2017learning} or recover actions from state transitions~\cite{torabi2018behavioral} for imitation learning, 
% (noisy, partially observable or unobservable). They are mainly rooted in the following two ideas. 1) 
% using only the states~\cite{merel2017learning} or  are two common techniques when the actions are unobservable or partially observable. 
% Two other common techniques to handle noisy actions are denoising through embedding/featuralization~\cite{duan2017one,wang2017robust,gao2018reinforcement} and explicitly modeling noises by learning multiple policies~\cite{li2017infogail,hausman2017multi}. 
%  For instance, the methods for handling noisy actions often assume full observability from actions, e.g.,~\cite{wang2017robust,duan2017one}, whereas the ones to handle unobservable actions completely ignore any information relevant to actions, e.g.,~\cite{merel2017learning,torabi2018behavioral}. On the other hand, the actions in demonstrations are often presented in mixed conditions, which may be unknown. For example,
% the conditions of actions in demonstrations may not be possibly known real demonstrations are 
% the noises and partial or none observations separately, which however often happed simultaneously.  
they rely solely on state information, and largely overlook the fact that a partial action sequence could still be available in one demonstration.  
% As for actions under mixed conditions, simply combining the existing methods could be challenging, if not impossible. 
It is thus necessary to design an algorithm that could handle demonstrations with partial action sequences. 

To this end, we propose a novel algorithm, Action-Guided Adversarial Imitation Learning (AGAIL), that can be applied to demonstrations with incomplete action sequences. The main idea of AGAIL algorithm is to divide the state-action pairs in demonstrations into state trajectories and action trajectories, and learns a policy from states with auxiliary guidance from actions, if available. To be more specific, AGAIL is built on \textit{adversarial imitation}, an idea of training a policy by competing it with a discriminator, which tries to distinguish between state-action pairs from expert as opposed to from the policy~\cite{ho2016generative}. AGAIL further divides the state-action matching into two components, state matching and action guidance, and simultaneously maintains three networks: a generator, a discriminator, and a guide, as shown in Figure~\ref{fig:framework}. The generator generates a policy via a state-of-the-art policy gradient method; the discriminator distinguishes the state distribution between demonstrations and the learned policy, and assigns rewards to the generator; and the guide provides additional credits by maximizing the mutual information between generated actions and demonstrated actions if available. The policy net and the state discrimination net are trained by competing with each other, while the action guidance net is trained only when actions for specific states are available. 
% This unified framework actions as an additional infomation and propose a new imitation learning algorithm, called Action Constrained Adversarial Imitation Learning (ACAIL). ACAIL learns an expert policy from observations with different action conditions: no actions, partial actions, noisy actions or noisy partial actions. Basically, it is a unified framework for general imitation learning tasks. ACAIL works as follows. First, it converts observations into features, i.e., states, in an unsupervised manner and then adopt Generative Adversarial Imitation Learning (GAIL), a recent breakthrough in imitation learning, to match the state visitations of the learned policy with that of the demonstrated state trajectories. Second, ACAIL treats the demonstrated actions as a side information, which is then incorporated into the learned policy by a causal entropy optimization. 
We present a theoretical analysis of AGAIL to show its correctness. Through various experiments on different levels of incompleteness of actions in demonstrations, we show that AGAIL consistently delivers comparable performance to two state-of-the-art algorithms even when the demonstrations provided are incomplete. 
% The proposed AGAIL algorithm can also be easily adapted for any existing IRL algorithms. \notice{The main contribution of our work is an algorithm that learns...}

% Imitation learning with policy gradients [Ho et al., 2016] is a recently proposed approach that uses gradient-based stochastic optimizers. Along with trust-region policy optimization (TRPO) [Schulman et al., 2015] as the optimizer, it is shown to be one of the most practical approaches that scales well to large-scale environments, i.e. high- dimensional state and action spaces. Generative adversarial imitation learning (GAIL) [Ho and Ermon, 2016], which is of our primary interest, is a recent instance of imitation learning algorithms with policy gradients. GAIL reformulates the imitation learning problem as a density matching problem, and makes use of generative adversarial networks (GANs) [Goodfellow et al., 2014]. This is achieved by generalizing the representation of the underlying cost function using neural networks, instead of restricting it to the class of linear functions for the sake of simpler optimization. As a result, the policy being learned becomes the generator, and the cost function becomes the discriminator. Based on the promising results from GAIL, a number of improvements appeared in the literature [Wang et al., 2017, Li et al., 2017]. 
\section{Related Work}
This section briefly introduces imitation learning algorithms, and then discusses how demonstrations with partial or unobservable actions are handled by previous studies. 

% \red{how to introduce the imitation learning? behavior cloning and inverse reinforcement learning}
% which typically fall under one of two broad categories: behavioral cloning (BC) and inverse reinforcement learning (IRL). 
% Imitation learning aims at mimicking expert behavior, which can be realized in multiple ways. 
To solve an imitation learning problem, one simple yet effective method is Behavior Cloning (BC)~\cite{bain1999framework}, a supervised learning approach that directly learns a mapping from states to actions from demonstrated data~\cite{ross2010efficient}. Though successfully applied to various applications, e.g., autonomous driving~\cite{bojarski2016end} and drone flying~\cite{daftry2016learning}, BC suffers greatly from the \textit{compounding error}, a situation where minor errors are compounded over time and finally induce a dramatically different state distribution~\cite{ross2011reduction}. Another approach, Inverse Reinforcement Learning (IRL)~\cite{ng2000algorithms}, aims at searching for a reward/cost function that could best explain the demonstrated behavior. Yet the function search is ill-posed as the demonstrated behavior could be induced by multiple reward/cost functions. Constraints are thereby imposed on the rewards or the policy to ensure the optimality uniqueness of the demonstrated behavior. For example, the reward function is usually defined to be a linear~\cite{ng2000algorithms,abbeel2004apprenticeship} or convex~\cite{syed2008apprenticeship} combination of the state features. The learned policy is also assumed to have the maximum entropy~\cite{ziebart2008maximum} or the maximum causal entropy~\cite{ziebart2010casual_entropy}. These explicit constraints, on the other hand, potentially limit the generability of the proposed methods~\cite{ho2016generative}. 
% Ng and Russell proposed to represent the reward as a linear combination of state features~\cite{ng2000algorithms}, which was later adapted to the feature matching by Abbeel and Ng~\cite{abbeel2004apprenticeship}. Similarly, Syed~\emph{et al.} assumed that the reward function is an unknown convex optimization of the basis reward functions~\cite{syed2008apprenticeship}. 
Only recently, Finn~\emph{et al.} have proposed to skip the reward constraints and used demonstrations as an implicit guidance for reward searching~\cite{finn2016guided}. 
% Furthermore, in order to ensure the uniqueness of the expert policy, some constraints, e.g., the maximum entropy principle~\cite{ziebart2008maximum} and the maximum causal entropy~\cite{Ziebart:2010:MIV:3104322.3104481}, could also be imposed on the policy. 
Nevertheless, the reward-based methods are computationally intensive and hence are limited to simple applications~\cite{ho2016generative}. To address this issue, Generative Adversarial Imitation Learning (GAIL)~\cite{ho2016generative} was proposed to use a discriminator to distinguish whether a state-action pair is from an expert or from the learned policy. Since GAIL has achieved state-of-the-art performance in many applications, we thus derive our algorithms based on the GAIL method. For more details on GAIL, refer to Prelminary.

The aforementioned algorithms, however, can hardly handle the demonstrations with partial or unobservable actions. One idea to learning from these demonstrations is to first recover actions from states and then adopt standard imitation learning algorithms to learn a policy from the recovered state-action pairs. For example, Torabi~\emph{et al.} recovered actions from states by learning a dynamic model of state transitions, and then use a BC algorithm to find the optimal policy~\cite{torabi2018behavioral}. However, the performance of this method is highly dependent on the learned dynamic model, and may fail when the states transit with noise. 
Instead, Merel~\emph{et al.} proposed to learn from only state (or state feature) trajectories. They extended the GAIL framework to learn a control policy from only states of motion capture demonstrations~\cite{merel2017learning}, and showed that partial state features without demonstrator actions suffice for adversarial imitation. 
% They pointed out that the original presentation of GAIL was restricted to imitation of single skills from complete state-action trajectories, where the demonstrator shared the same body and policy parameterization as the imitator. They further showed that: (a) partial state featurizations without demonstrator actions suffice for adversarial imitation; (b) the body structure and physical parameters (i.e. body dynamics) need not match between the demonstrator and the imitator; and (c) robust transitions between behaviors naturally emerge by training on multiple behaviors. However, Merel~\emph{et al.} presented these empirical results without giving any rigorous analysis.
Similarly, Eysenbach~\emph{et al.} pointed out that the policy should control which states the agent visits, and thus used states to train a policy by maximizing mutual information between the policy and the state trajectories~\cite{eysenbach2018diversity}.
% \textit{Diversity is All you need}, a novel framework for imitation learning without the definition of rewards
% , to maximize the mutual information between the skills and states. 
% To ensure that states, not actions, are used to distinguish skills, they minimized the mutual information between skills and action given the states.
% -- compress then use
Other studies have also tried to learn from raw observations, instead of states. For instance, Stadie~\emph{et al.} extracted features from observations by the domain adaptation method to ensure that experts and novices are in the same feature space~\cite{stadie2017third}. However, only using demonstrated states or state features may require a huge number of environmental interactions during the training since any possible information from actions is ignored.
\section{Preliminary}
An infinite-horizon, discounted Markov Decision Process (MDP) is modeled by tuple $(\mathcal{S} , \mathcal{A}, P, r , \rho_0, \gamma)$, where $\mathcal{S}$ is the state space, $\mathcal{A}$ is the action space, $P:\mathcal{S}\times\mathcal{A}\times\mathcal{S}\rightarrow \mathbb{R}$ denotes the state transition probability, $r:\mathcal{S}\times\mathcal{A}\rightarrow\mathbb{R}$ represents the reward function, $\rho_0:\mathcal{S} \rightarrow\mathcal{A}$ is the initial state distribution, and $\gamma\in(0, 1]$ is a discount factor. A stochastic policy $\pi\in\Pi$ is $\pi: \mathcal{S}\times\mathcal{A}\rightarrow [0, 1]$. Let $\tau_E$ denote a trajectory sampled from expert policy $\pi_E$: $\tau_E = \big[ (s_0, a_0), (s_1, a_1), ..., (s_n, a_n) \big]$. We also use $\tau_{Es}$ and $\tau_{Ea}$ to denote state component and action component in $\tau_E$: $\tau_E = (\tau_{Es}, \tau_{Ea})$, $\tau_{Es}= [ s_0, s_1, ..., s_n]$ and $\tau_{Ea}= [ a_0, a_1, ..., a_n]$. We use the expectation with respect to a policy $\pi$ to denote an expectation with respect to trajectories it generates: $\mathbb{E}_{\pi}\big[ r(s, a) \big]\triangleq \mathbb{E}\big[\sum_{t=0}^{\infty}\gamma^{t}r(s_t, a_t)\big]$, where $s_0\sim\rho_0$, $a_t\sim\pi(a_t|s_t)$, $s_{t+1}\sim P(s_{t+1}|a_t, s_t)$. 

To address the imitation learning problem, we adopt the apprenticeship learning formalism~\cite{abbeel2004apprenticeship}: the learner finds a policy $\pi$ that performs not worse than expert $\pi_E$ with respect to an unknown reward function $r(s, a)$. We define the occupancy measure $\rho_{\pi} \in\mathcal{D}:\mathcal{S}\times\mathcal{A}\rightarrow\mathbb{R}$ of a policy $\pi\in\Pi$ as: $\rho_{\pi}(s, a)=\pi(a|s)\sum_{t=0}^{\infty}\gamma^{t}p(s_t=s|\pi)$~\cite{puterman2014markov}. Owing to the one-to-one correspondence between $\Pi$ and $\mathcal{D}$, an imitation learning problem is equivalent to a matching problem between $\rho_{\pi}(s, a)$ and $\rho_{\pi_E}(s, a)$. A general objective of imitation learning is
\begin{equation}\label{equ:gail_objective}
\arg\min_{\pi \in \Pi} - \lambda_1 H(\pi) + \psi^{*}(\rho_{\pi} - \rho_{\pi_{E}}) 
\end{equation}
where $H(\pi)\triangleq \mathbb{E}_{\pi}\big[ -\log \pi(s, a) \big]$, is the $\gamma$-discounted causal entropy of the policy $\pi(s, a)$, and $\psi^{*}$ is a distance measure between $\rho_{\pi}$ and $\rho_{\pi_E}$. In GAIL framework, the distance measure is defined as follows:
\begin{equation}\label{equ:gail_relaxation}
\psi_{GA}^{*}(\rho_{\pi} - \rho_{\pi_{E}}) = \max_{D} \mathbb{E}_{\pi}\big[ \log D \big] + \mathbb{E}_{\pi_E}\big[ \log (1 - D) \big]
\end{equation} 
where $D\in(0, 1)^{\mathcal{S}\times\mathcal{A}}$ is a discriminator with respect to state-action pairs. Based on this formalism, imitation learning becomes training a generator against a discriminator: generator $\pi_{\theta}$ generates state-action pairs while the discriminator tries to distinguish them from demonstrations. The optimal policy is learned when the discriminator fails to draw a distinction.

\paragraph{Problem formulation.} We now formulate the problem of imitation learning from incomplete demonstrations. Without loss of generality, we define a demonstration to be incomplete based on the action condition: a demonstration $\tau_E$ is said to be incomplete if part(s) of its action component $\tau_{Ea}$ is missing, i.e., $|\tau_{Ea}| \leq |\tau_{Es}|$. Figure~\ref{fig:framework} illustrates $\tau_{Es}$ and $\tau_{Ea}$ in an incomplete demonstration. Then imitation learning from incomplete demonstrations becomes the learner finds a policy $\pi$ that performs not worse than the expert $\pi_E$, which is provided in state trajectory samples and action trajectory samples, i.e., $\tau_{Es} = \{ \tau_{Es}^i \}$, $\tau_{Ea} = \{ \tau_{Ea}^{i} \}$ and $|\tau_{Ea}^{i}| \leq |\tau_{Es}^{i}| \quad \forall i$. 

\section{Action-Guided Adversarial Imitation}
% This section formulates the problem of imitation learning from noisy and incomplete demonstrations, and then introduces the AGAIL algorithm. 
We now describe our imitation learning algorithm, AGAIL, which combines state-based adversarial imitation with action-guided regularization. Motivated by the studies on utilizing demonstrations to steer explorations in Reinforcement Learning~\cite{brys2015reinforcement,kang2018policy}, we propose to separate the demonstrations into two parts: state trajectories and action trajectories. The state trajectories $\tau_{Es}=\{\tau_{Es}^{i}\}$ are for learning an optimal policy, while the action trajectories $\tau_{Ea}=\{\tau_{Ea}^{i}\}$ provides auxiliary information to shape the learning process. AGAIL has two parts: a state-based adversarial imitation, and an action-guided regularization. The pseudo-code of AGAIL is given in Algorithm~\ref{alg:agail}.

\begin{algorithm}[tb]
\caption{Action-guided adversarial imitation learning}
\label{alg:agail}
\textbf{Input}: expert
trajectories $\tau_E = \{(\tau_{Es}^{i},\tau_{Ea}^{i})\} \sim \pi_E$
\\
\textbf{Parameter}: Policy, discriminator and posterior parameters $\theta_0$, $\omega_0$, $\psi_0$; hyperparameters $\alpha$ and $\beta$\\
\textbf{Output}: Learned policy $\pi_{\theta}$
\begin{algorithmic}%[1] enables line numbers
\FOR{$i=0, 1, 2, ...$}
\STATE Sample trajectories: $\tau^i \sim \pi_{\theta_i}$ during each rollout. 
\STATE Sample states $s^{i} \sim \tau_{s}^{i}$, $s_E^{i} \sim \{\tau_{Es}^{i}\}$ by same batch size.  
\STATE Update $\omega_i$ to $\omega_{i+1}$ for $D_{\omega}$ based on Equation~\ref{equ:d_gradient}.
\STATE Query $\{a_E^{i}\}$ and run $\pi_{\theta_i}$ on $\{s_E^{i}\}$ to collect $\{a^{i}\}$.
\STATE Update $\psi_i$ to $\psi_{i+1}$ for $Q_{\psi}$ based on Equation~\ref{equ:g_gradient}.
\STATE Update $\theta_i$ to $\theta_{i+1}$ via TRPO for Equation~\ref{equ:agail_objective} with rewards  $r(s, a) = \alpha D_{\omega_{i+1}}(s) + \beta Q_{\psi_{i+1}}(a_E | s, a) \quad a_E\sim\tau_{Ea}$
\ENDFOR
\end{algorithmic}
\end{algorithm}

\subsection{State-Based Adversarial Imitation}
% We start from the dual method of solving MDPs. Defining $\rho(s) = \sum_{a}\rho(s, a)$. Then the Bellman flow constraint becomes 
% \begin{equation*}
% \rho_{\pi}(s) = p_{0}(s) + \sum_{s^{\prime}} \gamma P(s|s^{\prime}) \rho_{\pi}(s^{\prime}) 
% \end{equation*}
% where $P(s|s^{\prime}) = \sum_{a} P(s|s^{\prime}, a)\pi(a|s^{\prime})$. 

% introduce the general form of an imitation learning
We start from the occupancy measure matching~\cite{littman1995complexity,ho2016generative} in imitation learning and show that a policy $\pi$ can be learned from state trajectories $\{\tau_{Es}\}$, which we called state-based adversarial imitation. 
% learning with states
In general, any imitation learning problem can be converted into a specific matching problem between two occupancy measures: one with respect to the expert policy, $\rho_{\pi_E}(s, a)$, and another with respect to the learned policy, $\rho_{\pi}(s, a)$~\cite{pomerleau1991efficient}. However, $\rho_{\pi_E}(s, a)$ cannot be calculated exactly since the expert demonstrations are only provided in the form of a finite set of trajectories. Thus the matching of two occupancy measures is further relaxed into a regularization as shown in Equation~\ref{equ:gail_objective}, with $\psi$ penalizes the difference between the two occupancy measures. It has been shown that many imitation learning algorithms, e.g., apprenticeship learning methods~\cite{abbeel2004apprenticeship,syed2008apprenticeship}, are actually originated from some specific variant of this regularizer~\cite{ho2016generative}. Hence, we derive our algorithm based on Equation~\ref{equ:gail_objective}.

To optimize Equation~\ref{equ:gail_objective}, both states and actions need to be available in demonstrations, especially for the second term $\psi^{*}(\rho_{\pi} - \rho_{\pi_E})$ (the first term is constant if we define the policy to be Gaussian). Ho and Ermon have demonstrated that, if we choose the $\psi^{*}$ to be $\psi_{GA}^{*}$ in Equation~\ref{equ:gail_relaxation}, then $D\in (0, 1)$ relies only on rewards $r(s, a)$, and can be defined as a special function of $(s, a)$~\cite{ho2016generative}. Thus, after choosing $\psi$, the definition of $r$ determines the form of $D$. In many practical applications, the reward $r$ is defined based solely on states. For example, when training a human skeleton to walk in a simulation environment, the reward is defined mainly on the body positions and velocities, i.e., states. This is partly because the observed state trajectories are sufficiently invariant across a human skeleton~\cite{merel2017learning}.

We now show that $\psi^{*}(\rho_{\pi} - \rho_{\pi_E})$ can be approximated by another distance measure that is defined only on states. Assuming the reward $r$ is defined (mainly) on states $s$ and $\psi^*=\psi_{GA}^*$, we can now define $D$ as $D(s)\in(0, 1)$, a function with respect to states only. Let $\nu(s)$ denote the state visitations $\nu(s)=\sum_{t=0}^{\infty}\gamma^{t}p(s_t=s|\pi)$. Accordingly, the occupancy measure $\rho_{\pi}(s, a)$ can be written as $\rho_{\pi}(s, a) = \pi(a|s) \nu_{\pi}(s)$. Equation~\ref{equ:gail_relaxation} now becomes
\begin{align}\label{equ:agail_d_objective}
& \max_{D\in (0, 1)} \mathbb{E}_{\pi}\big[ \log D(s, a) \big] + \mathbb{E}_{\pi_E}\big[ \log (1 - D(s, a)) \big] \nonumber \\
\approx & \max_{D\in (0, 1)} \mathbb{E}_{\pi}\big[ \log D(s) \big] + \mathbb{E}_{\pi_E}\big[ \log (1 - D(s)) \big] \nonumber \\
= & \sum_{s, a} \max_{D} \rho_{\pi}(s, a) \log D(s) + \rho_{\pi_E}(s, a) \log (1- D(s)) \nonumber \\
= & \sum_{s, a} \max_{D} \resizebox{.80\linewidth}{!}{$\pi(a|s) \nu_{\pi}(s) \log D(s) + \pi_E(a|s) \nu_{E}(s) \log (1- D(s))$} \nonumber \\
% = & \sum_{s, a} \max_{D} \pi(a|s) \nu_{\pi}(s) \log D(s) + \pi_E(a|s) \nu_{E}(s) \log (1- D(s)) \nonumber \\
= & \sum_{s} \max_{D\in (0, 1)} \nu_{\pi}(s) \log D(s) + \nu_{E}(s) \log (1- D(s)) \nonumber \\
= & \max_{D} \mathbb{E}_{s\sim\pi} \log D(s) + \mathbb{E}_{s\sim\pi_E}\log (1- D(s))
\end{align}

% Unlike Equation~\ref{equ:gail_relaxation}, the above equation relies only on state samples: $D$ is a function with respect to states, and the expectation of policies $\pi$ and $\pi_E$ is also over state samples. 
This equation implies that, rather than matching the distribution of state-action pairs, we can instead compare the state distribution with the demonstrations to train an optimal policy. Similar to GAIL framework, we train a discriminator $D(s)$ to distinguish the state distribution between the generator and the true data. When $D(s)$ cannot distinguish the generated data from the true data, then $\pi$ has successfully matched the true data. In this setting, the learner's state visitations $\nu_{\pi}(s)$ is analogous to the data distribution from the generator, and the expert's state visitations $\nu_{\pi_E}$ is analogous to the true data distribution. We now introduce a discriminator network $D_{\omega}:\mathcal{S}\rightarrow(0, 1)$, with weights $\omega$, and update it on $\omega$ to maximize Equation~\ref{equ:agail_d_objective} with the following gradient.
\begin{equation}\label{equ:d_gradient}
\mathbb{E}_{s}\big[ \nabla_{\omega_i} \log D_{w_i}(s) \big] + \mathbb{E}_{s_E}\big[ \nabla_{\omega_i} \log (1- D_{w_i}(s)) \big]
\end{equation}
We also parametrize the policy $\pi$, i.e., the generator, with weight $\theta$, and optimize it with Trust Region Policy Optimization (TRPO)~\cite{schulman2015trust} as it changes the policy $\pi_{\theta}$ within small trust region to avoid policy collapse. The generator $\pi_{\theta}$ and the discriminator $D_{\omega}(s)$ forms the structure of state-based adversarial imitation.

\subsection{Action-Guided Regularization}
% why need partial actions
One downside of the state-based adversarial imitation described above is the lack of considering any available actions in demonstrations. Although incomplete and partially available, these action sequences can still provide useful information for the policy learning and explorations~\cite{kang2018policy}. We now considers how to utilize the partial actions in demonstrations. One technique that is widely adopted in Learning from Demonstration is reward shaping~\cite{ng1999policy,brys2015reinforcement}, i.e., defining potentials for demonstrated actions to modify rewards. However, the definition of an appropriate potential function for demonstrated actions is non-trivial, especially when the actions are continuous and high-dimensional. We instead borrow the idea from InfoGAN~\cite{chen2016infogan} and InfoGAIL~\cite{li2017infogail} to incorporate demonstrated actions into learning process by information theories.
% treating actions as labels (e.g., in BC algorithms~\cite{bain1999framework,torabi2018behavioral}) or optimal behavior to imitate (e.g., in IRL algorithms~\cite{ng2000algorithms,abbeel2004apprenticeship}), we treat the demonstrated actions (if available) as auxiliary information, which can be used to guide the training process. The same strategy has also been successfully applied when the demonstrations are sub-optimal. 
In particular, there should be high mutual information between two distributions: the demonstrated actions $a_E$ and the generated actions $a\sim\pi(s_E)$ for any specific state $s_E$ that corresponds to the demonstrated actions. 
% Thus $I(a_E;\pi(a|s_E))$ should be high. 
In information theory, mutual information between $a_E$ and $a\sim\pi(s_E)$, $I(a_E;a\sim\pi(s_E))$, measures the ``amount of information'' provided to $a_E$ when knowing $a\sim\pi(s_E)$. In other words, $I(a_E; a\sim\pi(s_E))$ is the reduction of uncertainties in $a_E$ when $a\sim\pi(s_E)$ is observed. 
% If $a_E$ and $\pi(a|s_E)$ are independent, then $I(a_E;Y)=0$, because knowing one variable reveals nothing about the other; by contrast, if $X$ and $Y$ are related by a deterministic, invertible function, then maximal mutual information is attained. 
Thus, we formulate an additional regularizer for the training objective: given any $a_E \in \{\tau_{Ea}\}$, we want $I(a_E|a\sim\pi(s_E)$ to have maximum mutual information, where $s_E$ is the state where the action $a_E$ is demonstrated, and $a$ is sampled from $\pi(s_E)$.
 
% incorporate the mutula information into the learning process
% In particular, the regularization seeks to maximize the mutual information between demonstrated actions and actions predicted by the learned policy at demonstrated states, denoted as $I(a_E; \pi(s))$. 

However, the mutual information is hard to maximize as it requires the posterior $P(a_E|a\sim\pi(s_E))$. We adopt the same idea in InfoGAIL to introduce a variational lower bound, $L_I(\pi, Q)$, of the mutual information $I(a_E;a\sim\pi(s_E))$:
\begin{align*}
L_I(\pi, Q) &= \mathbb{E}_{a_E\sim\{\tau_{Ea}\}}\big[ \log Q(a_E|a, s_E) \big] + H(a_E) \\
&\leq I\big(a_E; a\sim\pi(s_E) \big)
\end{align*}
where $Q(a_E|a, s_E)$ is an approximation of the true posterior $P(a_E|a\sim\pi(s_E))$. We parameterize the posterior approximation $Q$ with weights $\psi$, i.e., $Q_{\psi}(a_E|a, s_E)$, by a neural network and update $Q_{\psi}$ by the following gradients:
\begin{equation}\label{equ:g_gradient}
-\mathbb{E}_{s_E, a_E}\big[ \nabla_{\psi_i} \log Q_{\psi_i}(a_E|a, s_E) \big]
\end{equation}
Note that the mutual information is maximized between the distribution of demonstrated actions and the distribution of generated actions from the same state. The weights of $Q$ are shared across all demonstrated actions and states.

\begin{table*}[ht]
\centering
\begin{tabular}{rcccccccc}
\toprule
\multicolumn{1}{c}{\multirow{2}{*}{\textbf{Env.}}}
& \multicolumn{1}{c}{\multirow{2}{*}{$\mathcal{S}\times\mathcal{A}$}} 
& \multicolumn{6}{c}{\textbf{Empirical Return}} \\ 
\multicolumn{1}{c}{}&
\multicolumn{1}{c}{}&
\multicolumn{1}{c}{\textit{TRPO}} &
\multicolumn{1}{c}{\textit{GAIL}} &
\multicolumn{1}{c}{\textit{State}} &
\multicolumn{1}{c}{\textit{AGAIL}$_{.00}$} &
\multicolumn{1}{c}{\textit{AGAIL}$_{.25}$} &
\multicolumn{1}{c}{\textit{AGAIL}$_{.50}$} &
\multicolumn{1}{c}{\textit{AGAIL}$_{.75}$} \\
\midrule
CartPole & $\mathbb{R}^4\times\{0, 1\}$ & 196.4$^{\pm2}$ & 188.6$^{\pm6}$ & 188.3$^{\pm8}$ & 18.4$^{\pm1}$ & 197.2$^{\pm1}$ & 193.6$^{\pm5}$ & 197.9$^{\pm1}$ \\
Hopper  & $\mathbb{R}^{11}\times\mathbb{R}^{3}$& 2.6e3$^{\pm96}$ & 2.5e3$^{\pm181}$ & 2.6e3$^{\pm203}$ & 1.0e3$^{\pm23}$ & 1.4e3$^{\pm269}$ & 1.5e3$^{\pm309}$ & 2.7e3$^{\pm131}$ \\ 
Walker2d & $\mathbb{R}^{17}\times\mathbb{R}^{6}$& 2.4e3$^{\pm180}$ & 2.3e3$^{\pm280}$ & 2.0e3$^{\pm121}$ & 2.3e3$^{\pm84}$ & 2.6e3$^{\pm150}$ & 2.3e3$^{\pm109}$ & 2.2e3$^{\pm200}$ \\
Humanoid & $\mathbb{R}^{376}\times\mathbb{R}^{17}$& 523.9$^{\pm8}$ & 509.2$^{\pm14}$ & 544.7$^{\pm12}$ & 586.4$^{\pm14}$ & 571.3$^{\pm10}$ & 548.6$^{\pm12}$ & 542.3$^{\pm6}$ \\
\bottomrule
\end{tabular}
\caption{Environment specifications and  numerical results. }
\label{tab:result_data}
\end{table*}

\begin{figure*}
    \centering
    \includegraphics[width=1.0\linewidth]{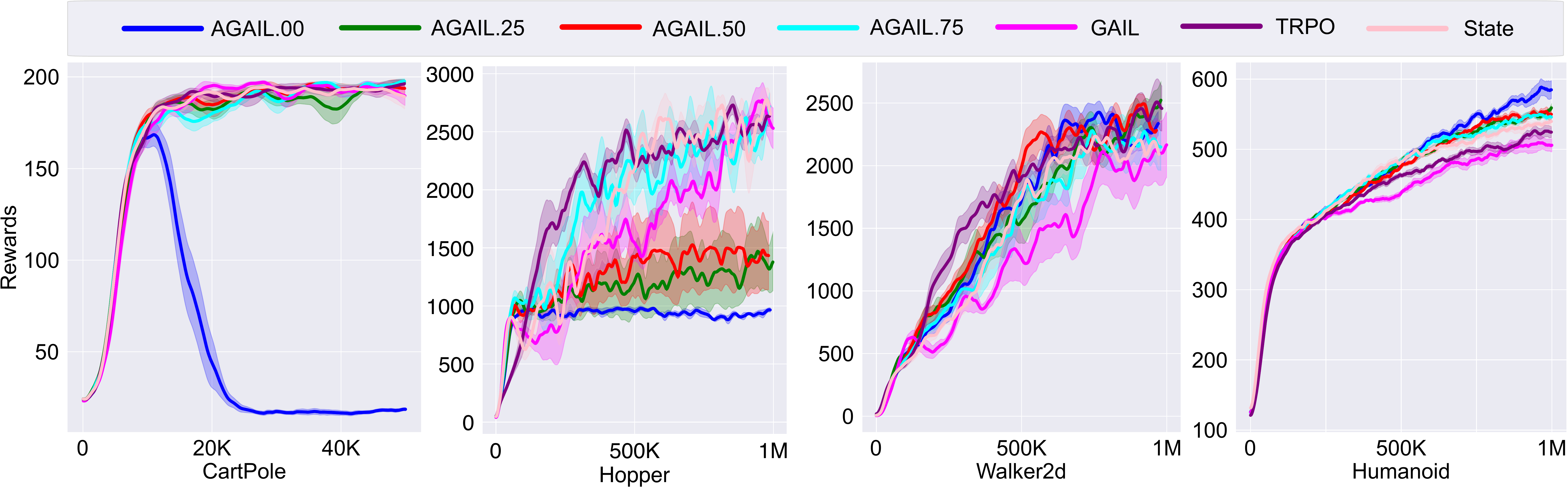}
    \caption{Reward curves of AGAIL\{.00, 0.25, 0.50, .75\}, TRPO, GAIL and state-GAIL (\{.xx\} denotes the incompleteness ratio).} 
    \label{fig:result_effectiveness}
\end{figure*}

Now, we present the Action-Guided Adversarial Imitation Learning (AGAIL) algorithm. The learning objective that combines the state-based adversarial imitation and the action-guided regularization is:
\begin{align}\label{equ:agail_objective}
\min_{\pi \in \Pi} \big[&-\lambda_1 H(\pi_{\theta}) -\lambda_2 L_I(\pi_{\theta}, Q_{\psi}) \nonumber + \\
&\max_{D} \mathbb{E}_{s\sim\pi_{\theta}} \log D_{\omega} + \mathbb{E}_{s\sim\pi_E}\log (1- D_{\omega}) \big]
\end{align}
where $\lambda_1, \lambda_2 > 0 $ are two hyperparameters for the casual entropy of policy $\pi_{\theta}$ and the mutual information maximization respectively. Optimizing the objective involves three steps:  maximizing Equation~\ref{equ:d_gradient}, minimizing Equation~\ref{equ:g_gradient}, and minimizing Equation~\ref{equ:agail_objective} with fixed $D_{\omega_i}$ and $Q_{\psi_i}$. The first step is similar as GAIL. In second step, we assume that all demonstrated state-action pairs $(s_E^i, a_E^i)$ are independent and only update $Q_{\psi}$ when $a_E$ is available for $s_E$. When updating $Q_{\psi}$, we use $(a_E^i, a, s_E^i)$, where $a\sim \pi_{\theta}(s_E)$; when using $Q_{\psi}$ as additional rewards for $(s, a)$, we sample $a_E\sim\tau_{Ea}$ and then feed a tuple $(a_E, a, s)$ to $Q$. To conduct the third-step optimization, we use both $D(s)$ and $Q(a_E^i|a, s_E^i)$ as rewards to update $\pi_{\theta}$ on state $s_E$, i.e., $r(s_E, a) = \alpha D_{\omega}(s_E) + \beta Q(a_E^i | s, a)$ where $\alpha$ and $\beta$ are coefficients. In the experiment, we set $\alpha$ to 1 and relate $\beta$ to the incompleteness ratio $\eta\in (0, 1)$ of actions in demonstrations, $\beta = 1 - \eta$. The three steps are run iteratively until convergence. An outline for this procedure is given in Algorithm~\ref{alg:agail}.

\section{Experiment}

We want to investigate two aspects of AGAIL: the effectiveness of learning from incomplete demonstrations, and the robustness when the degree of incompleteness changes. Specifically, we compare AGAIL to three algorithms, TRPO, GAIL and state-only GAIL, to show its learning performance. The reason for choosing TRPO is that, given true reward signals, TRPO delivers the state-of-the-art performance, which can then be referred to as the ``expert'' when the true rewards are unknown. We select GAIL as it is the state-of-the-art for imitation learning when demonstrations are complete. We also adopt state-GAIL~\cite{merel2017learning} (using states only to train GAIL and equivalent to AGAIL.100) to show the performance boost introduced by action guidance. 
% Note that the state-only method in~\cite{torabi2018behavioral} is inapplicable as calculating maximum-likelihood of an action given two states is computationally infeasible in very high dimensional environment, e.g., Humanoid needs to deal with 2.4M (376*376*17) continuous dimensions. 
The characteristics of each algorithm are listed below:
\begin{compactitem}
\item \textbf{TRPO}: true $r$; no $\tau_{Es}$ and no $\tau_{Ea}$
\item \textbf{GAIL}: discriminator $r$; $\tau_{Es}$ and $\tau_{Ea}$
\item \textbf{State-GAIL}: discriminator $r$; $\tau_{Es}$ and no $\tau_{Ea}$
\item \textbf{AGAIL}: discriminator \& guide $r$; $\tau_{Es}$ and partial $\tau_{Ea}$
\end{compactitem}
% \noindent -- TRPO: true $r$; no $\tau_E$\\
% \noindent -- GAIL: discriminator $r$; $\tau_{Es}$ and $\tau_{Ea}$\\
% \noindent -- State GAIL: discriminator $r$; $\tau_{Es}$\\
% \noindent -- AGAIL: discriminator \& guide $r$; $\tau_{Es}$ and partial $\tau_{Ea}$\\
In addition, we vary the level of incompleteness of demonstrations to showcase the robustness of AGAIL. Four simulation tasks, Cart Pole, Hopper, Walker and Humanoid (from low-dimensional to high-dimensional controls), are selected to cover discrete and continuous state/action space, and the specifications are listed in Table~\ref{tab:result_data}. Note that the rewards defined in all four environments are mainly dependent on the states. For example, the rewards for Cart Pole is set as a function of positions and angles of the pole; the rewards for Hopper, Walker and Humanoid all have a significant weight on states~\cite{brockman2016openai}. Thus our assumption that the reward $r$ is (mainly) a function of the state $s$ holds for all experimental environments. 

\begin{figure*}
    \centering
    \includegraphics[width=1.0\linewidth]{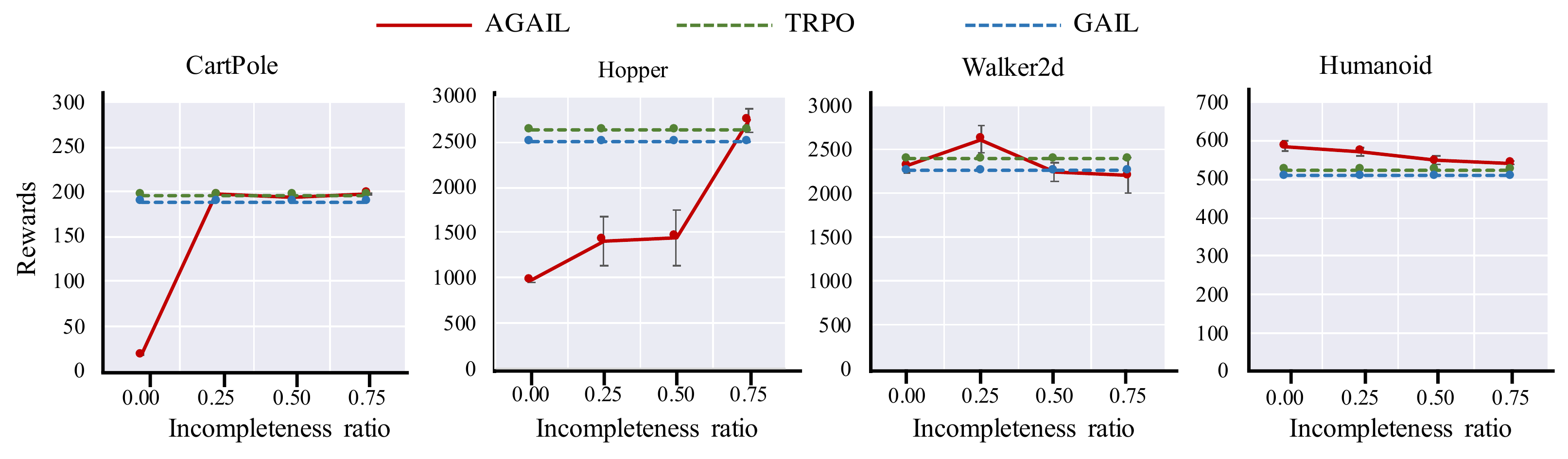}
    \caption{AGAIL performance versus TRPO and GAIL baselines as the incompleteness ratio changes}
    \label{fig:result_robustness}
\end{figure*}

% implementation details
\paragraph{Implementations.} \blue{We use stochastic policy parametrized by three fully connected layers (100 hidden units and Tanh activation), and construct the value network by sharing the layers with the policy network. Both policy net and  value net are optimized through gradient descend with Adam optimizer. Demonstrations are collected by running a policy trained via TRPO. We then randomly mask out actions to manipulate the incompleteness with four ratios (0\%, 25\%, 50\%, and 75\%): 0\% means all the actions are available while 75\% means 75\% of the actions in each demonstration are masked out. All experiments are run for six times with different initialization seeds (0-5). We use empirical returns to evaluate performance for the learned policy. All algorithms\footnote{See project page: \url{https://mingfeisun.github.io/agail/}} are implemented based on the work~\cite{brockman2016openai}. }

% Please add the following required packages to your document preamble:
% \usepackage{multirow}

% \paragraph{Experimental results.} 
\subsubsection{Experiment Results} 
\blue{We first compare the performance of AGAIL with TRPO, GAIL and state-GAIL in multiple control tasks. The average accumulated rewards are given in Table~\ref{tab:result_data} and the learning curves are plotted in Figure~\ref{fig:result_effectiveness}. The numerical results in Table~\ref{tab:result_data} show that AGAIL algorithm achieves learning performance comparable with that of TRPO (true rewards) and GAIL (complete demonstrations), and outperforms state-GAIL. Specifically, in CartPole tasks, AGAIL\{.25, .50, .75\} all achieve almost the same performance as that of TRPO and GAIL, even if it is trained with incomplete actions. The same phenomenon is observed in Walker2d and Humanoid environments. We also notice that, AGAIL\{.00, .25, .50, .75\} all outperform state-GAIL in Walker2d and Humanoid. Such performance boost in AGAIL, especially in Humanoid, further shows that the guidance layer is vital for AGAIL. However, in contrast to Walker2d and Humanoid, AGAIL.00 performs poorly in CartPole and Hopper. Such performance drop in CartPole and Hopper may possibly be caused by the qualities of demonstrations, i.e., the extent to whether demonstrations are good samples to show the expected optimal behaviour in expert policy~\cite{brys2015reinforcement}. The TRPO policy of these tasks (especially Hopper), though delivering good results in general, suffers from big performance fluctuations. Any one of the checkpoints from the TRPO policy could be impaired by the fluctuations regardless of its returns. In our experiment, demonstrations are generated by running one selected checkpoint (e.g., the one with the highest return) out of all possible TRPO checkpoints, which may overfit one batch of examples and produce actions that fail to scale. Forcefully requiring the policy actions to share similar distribution of these actions could thus lead to policy collapse.}
% In our experiment, demonstrations are generated by running one checkpoint (e.g., one with the highest return) out of all TRPO checkpoints. The TRPO policy, though delivering good results in general, suffers from big performance fluctuations, especially in Hopper. One checkpoint from the TRPO policy, even if with the biggest returns, could be impaired by the fluctuations. For example, the selected checkpoint in Hopper may overfit one batch and produces actions that fail to scale. Thus forcefully requiring the policy actions to share similar distribution of these actions could lead to policy collapse. }
% \blue{However, once the demonstrations are sampled from a stable training checkpoint, such as TRPO training in Humanoid, the AGAIL consistently achieves better results than the other methods do, as shown in Figure 2.}
% The learning curves in Fig. 2 demonstrate that both TRPO and DQfD fail to explore sufficiently to obtain informative feedback from the sparse environments. The demonstration data insufficiency severely limits the learning ability of DQfD which usually requires as many demonstration data as self-generated ones. Furthermore, GAIL succeeds in CartPole but converges to the imperfect demonstration data in MontainCar. Our POfD outperforms GAIL by a significant margin in terms of both convergence rate and final performance.

We are surprised that the AGAIL, trained with incomplete demonstrations, e.g., AGAIL.75, even outperforms GAIL with a noticeable margin in Hopper, Walker2d and Humanoid. Meanwhile, AGAIL\{.00, .25, .50\} all performs worse than AGAIL.75, especially in Hopper. We also notice that, in the same environment, GAIL fails to deliver satisfying results across all tasks. GAIL, AGAIL\{.00, .25, .50\} are all trained with a large portion ($\geq 50\%$) of demonstrated actions, while AGAIL.75 and TRPO are trained with much less or no actions. \blue{One might wonder why incorporating more actions fail to improve performance. A possible explanation is that demonstrations are limited samples from a training checkpoint (e.g., the one with the highest returns) of an expert policy~\cite{ho2016generative}. If the checkpoint itself is from an unstable training process, e.g., TRPO training in Hopper, more demonstrations are likely to introduce more undesirable variances in action distributions~\cite{kang2018policy}, which consequently interferes with policy deriving~\cite{ross2011reduction}. The same phenomenon has been observed in~\cite{ho2016generative,baram2017end}. In contrast, if demonstrations are sampled by a checkpoint from a stable training, e.g., TRPO training in Humanoid, employing more actions could lead to better results. As shown in Figure~\ref{fig:result_effectiveness} Humanoid, AGAIL performance improves as more actions are utilized.} Further, results in Figure~\ref{fig:result_effectiveness} Hopper suggest that demonstrations, or more specifically the actions, are not helpful for agents to learn a policy. This highlights the importance of demonstration qualities and the necessity of algorithms to handle incomplete actions.

% Overall, Figure~\ref{fig:result_effectiveness} shows the proposed algorithm, AGAIL, is very effective in handling incomplete demonstrations as it provides comparable performance with two state-of-the-art algorithms, TRPO and GAIL. In Hopper and Humanoid environments, the AGAIL even outperforms all other algorithms. 

We then test the robustness of AGAIL. Figure~\ref{fig:result_robustness} shows how the AGAIL performance changes as the incompleteness ratio increases. We notice that in Hopper and Humanoid, AGAIL consistently obtains more returns than GAIL under different ratios of action incompleteness. It even achieves the highest returns when used to train the Humanoid. However, in Walker2d environment, the returns of AGAIL fluctuate widely. This may possibly be caused by the large variance during the training, as shown in the AGAIL training curves in Walker2d in Figure~\ref{fig:result_effectiveness}. In all four subfigures, the TRPO algorithm performs stably better than the GAIl. In Hopper environment, the TRPO obtains much higher returns than the GAIL, while, in other environments, they achieve comparable returns. This may further verify the above guess that the demonstrated actions for Hopper are largely suboptimal. 

% Overall, Figure~\ref{fig:result_robustness} shows that although with some fluctuations, the AGAIL algorithm could still provide comparable performance to TRPO and GAIL, even if the incompleteness ratio increases.  

Combining above discussions, we conclude that AGAIL is effective in learning from incomplete demonstrations, and consistently delivers robust performance under different incompleteness ratios of demonstrated actions.

\section{Conclusions}
We considered imitation learning from demonstrations with incomplete action sequences, and proposed a novel and robust algorithm, AGAIL, to learn a policy from incomplete demonstrations. AGAIL treats states and actions in demonstrations separately. It first uses state trajectories to train a classifier and a discriminator: the classifier tries to distinguish the state distributions of expert demonstrations from the state distributions of generated samples; the discriminator leverages the feedback from the clsssifier to train a policy. Meanwhile, AGAIL also trains a guide to maximize the mutual information between any demonstrated actions, if available, and the policy actions, and assigns additional rewards to the generator. Experiment results suggest that AGAIL consistently delivers comparable performance to the TRPO and GAIL even if trained with incomplete demonstrations. 
% Furthermore, the AGAIL algorithm improves the policy by leveraging the benefits of demonstration for exploration. A simple dynamic reward reshaping based optimization algorithm for POfD was provided that connects to the generative adversarial training and can be applied efficiently. The POfD was shown to be effective in encouraging the agent to explore around the nearby region of the expert policy and learning better policies, through extensive experimental results. To our best knowledge, POfD is the first one that can acquire knowledge from few and imperfect demonstration data to aid exploration in environments with sparse feedback.
% Acknowledgements should only appear in the accepted version.
\section*{Acknowledgements}

% \textbf{Do not} include acknowledgements in the initial version of the paper submitted for blind review.

% If a paper is accepted, the final camera-ready version can (and probably should) include acknowledgements. In this case, please place such acknowledgements in an unnumbered section at the  end of the paper. 

The project is sponsored by Innovation and Technology Fund (ITF) with No. ITS/319/16FP, and the National Key Research and Development Plan Grant No. 2016YFB1001200. 

% to colleagues who contributed to the ideas, and to funding agencies and corporate sponsors that provided financial support.

\newpage
\bibliographystyle{named}
\bibliography{bib}

\end{document}